\pdfoutput=1
\documentclass[11pt]{article}
\usepackage{amssymb}
\usepackage{utils/ACL2023}
\usepackage{times}
\usepackage{latexsym}
\usepackage[T1]{fontenc}
\usepackage[utf8]{inputenc}
\usepackage{microtype}
\usepackage{inconsolata}
\usepackage{multirow}
\usepackage{microtype}
\usepackage{graphicx}
\usepackage{graphicx,subcaption,ragged2e}
\usepackage[shortlabels]{enumitem}
\title{Hate Speech Targets Detection in Parler using BERT}

\author{Nadav Schneider \\
  Department of Electrical and \\ Computer Engineering, \\Ben-Gurion University, Israel \\
  \texttt{nadavsch@post.bgu.ac.il} \\\And
  Shimon Shouei  \\
  Department of Software and \\ Information Engineering, \\Ben-Gurion University, Israel \\
  \texttt{shouei@post.bgu.ac.il} \\\AND
  Saleem Ghantous  \\
  Department of Software and \\ Information Engineering, \\Ben-Gurion University, Israel\\
  \texttt{saleem@post.bgu.ac.il} \\\And
  Elad Feldman  \\
  Department of Software and \\ Information Engineering, \\Ben-Gurion University, Israel \\
  \texttt{eladfeld@post.bgu.ac.il}}

\begin{document}
\maketitle
\begin{abstract}
Online social networks have become a fundamental component of our everyday life. Unfortunately, these platforms are also a stage for hate speech. Popular social networks have regularized rules against hate speech. Consequently, social networks like Parler and Gab advocating and claiming to be free speech platforms have evolved. These platforms have become a district for hate speech against diverse targets. We present in our paper a pipeline for detecting hate speech and its targets and use it for creating Parler hate targets distribution. The pipeline consists of two models; one for hate speech detection and the second for target classification, both based on BERT with Back-Translation and data pre-processing for improved results. \\The source code used in this work, as well as other relevant sources, are available at: \textcolor{blue}{\url{https://github.com/NadavSc/HateRecognition.git}}.
\end{abstract}

\section{Introduction}
Online social networks have become a fundamental component of our everyday life. Unfortunately, these platforms are also a stage for hate speech. Popular social networks have regularized rules against hate speech. Consequently, social networks like Parler and Gab advocating and claiming to be free speech platforms have evolved. These platforms have become a district for hate speech \cite{DBLP}, \cite{israeli2022free} against diverse targets. We present in our paper a pipeline for detecting hate speech and its targets and use it for creating Parler hate speech targets distribution. Hate speech is defined as hate towards minorities. This work focuses on four main minorities; People of color, Muslims, Jews, and LGBT. The pipeline consists of two models; One for hate speech detection and the second for target classification, both based on BERT with Back-Translation and data pre-processing for improved results.

\section{Related Work}
Hate speech detection in social media has been a growing area of research in recent years. There have been numerous studies aimed at identifying hate speech and determining the targets of hate, and several machine learning approaches have been proposed to address this challenge. In this section, we provide a brief overview of some of the most relevant works in this area.

\cite{davidson2017automated} presents a machine learning approach to investigate the distinction between hate speech and offensive language in tweets using a crowd-sourced hate speech lexicon and a multi-class classifier. The results show that accurately separating hate speech from offensive language is challenging, and future work should consider the different uses of hate speech and the people who use it.

\cite{ibrohim2019multi} presents multi-label text classification for abusive language and hate speech detection on Indonesian Twitter. They used three machine learning classifiers and several data transformation methods with term frequency, orthography, and lexicon features. They showed fine results in detecting hate speech, but the performance was low for identifying the target, category, and level of hate speech.

% \cite{al2019detection} provides an overview of the problem of hate speech in social media, especially in the context of the Arabic language. It discusses the concept of cyber hate and differentiates it from other anti-social behavior, such as cyberbullying, abusive and offensive language, radicalization, and hate speech. They also review the recent contributions on hate speech and its related anti-social behavior topics and present challenges and recommendations for the Arabic hate speech detection problem.

\cite{lemmens2021improving} studies the effectiveness of using hateful metaphors as features for identifying the type and target of hate speech in Dutch Facebook comments. The results of experiments using SVM and BERT/RoBERTa models showed that using hateful metaphors as features improved model performance in both type and target prediction tasks. The study is significant as it provides insight into the challenges posed by implicit forms of hate speech, which are often difficult to detect. The results also highlight the importance of considering various information encoding methods when using hateful metaphors as features for hate speech detection.

\cite{zampieri2019predicting} presents a new dataset for offensive language identification called the Offensive Language Identification Dataset (OLID) and compares the performance of different machine learning models on it. This dataset contains tweets annotated for offensive content. The annotations in OLID include the type and target of offensive messages, making it different from previous datasets that only focused on specific types of offensive content, such as hate speech or cyberbullying. The article discusses the similarities and differences between OLID and previous datasets and conducts baseline experiments using support vector machines and neural networks. The results show that identifying and categorizing offensive language in social media is challenging but doable.

\section{Data Sets}
\label{dataset}
For this project, various data sets were evaluated for their suitability in training and validation. The data sets were annotated and selected based on their relevance to the problem being addressed. In particular, an annotated Parler data set \cite{israeli-tsur-2022-free} has been utilized for the hate speech detection task, while HateXplain \cite{mathew2021hatexplain} and Dialoconan \cite{Dialoconan} have been used for the target detection task. These two data sets were utilized for training, while Toxigen \cite{hartvigsen-etal-2022-toxigen} data set and a new Target Annotated Parler (TAP) data set we have annotated (Section \ref{target_annotated_parler}) have been used for evaluation. In the following section, a detailed description of these data sets will be provided

\subsection{Annotated Parler}
\label{annotated_parler}
The annotated Parler data set contains 10,121 posts from Parler with four labeled parameters for each; label mean in 1-5 scale, user id and a boolean disputable parameter. Assuming three is the threshold between hateful and non-hateful posts, the data consists of 31.85\% hateful posts and 68.15\% non-hateful. Most of the posts are political and many of them scored with a high label mean, although hate speech was defined as hate against minorities. This distinction and its effects will be discussed forwards.

\subsection{Target Annotated Parler (TAP)}
\label{target_annotated_parler}
We created a new data set to evaluate the target classification model's performance in our suggested pipeline. To create Target Annotated Parler dataset, we have annotated 276 Parler posts by classifying them into six classes: Jewish, Islam, Homosexual, African, Politician, and Other. The criteria for the classification followed the definitions mentioned in the literature. As mentioned, hate speech is only against minorities, and therefore, the classes Politician and Other are necessary to separate toxic text from hate speech against minorities.

Posts with hate level of five were taken randomly from Annotated Parler data set. Most of the hate posts were labeled as Politicians and Other, meaning no specific ethnic/sexual orientation group was targeted. Between ethnic and sexual orientation, most posts were targeted as Islam. The other most targeted groups were Africans or of African descent, then Jewish, and lastly, Homosexual.

In case of controversy regarding the labeling of a certain post, an open discussion was made until a final consensus was reached.

\subsection{HateXplain}\label{HateXplain}
The HateXplain benchmark data set contains approximately 20K posts from Twitter and Gab, with a target and hate label added through Amazon Mechanical Turk (MTurk). The data includes approximately 6K posts containing hate speech against various community groups, with the focus being on Islam, Jews, Black people, and LGBT, which are the four largest groups. The remaining groups were categorized as "Other." Each post in the dataset was annotated three times, and the final label was determined through majority voting.

\subsection{Dialoconan}\label{Dialoconan}
The DIALOCONAN dataset is a comprehensive collection of conversations between online haters and non-governmental organization (NGO) operators. It contains a total of 3059 dialogues with either 4, 6, or 8 turns, totaling 16625 turns. The dataset covers six main targets of hate, including Jews, the LGBT community, migrants, Muslims, people of color, and women. The data was obtained through human expert intervention and machine-generated dialogues, utilizing 19 strategies. The dataset serves as an invaluable resource for the study of hate speech and its effects.

\subsection{Toxigen}\label{Toxigen}
Toxigen is a large-scale and machine-generated dataset of 274,186 toxic and benign statements about 13 minority groups.
This dataset uses a demonstration-based prompting framework and an adversarial classifier-in-the-loop decoding method to generate subtly toxic and benign text with a massive pre-trained language model (GPT-3). Controlling machine generation in this way allows Toxigen to cover implicitly toxic text at a larger scale, and about more demographic groups than previous resources of human-written text. Toxigen can be used to fight human-written and machine-generated toxicity.
Two Toxigen's variations have been used \ref{Toxigen} to evaluate the target classification model on a different data set: 
\begin{enumerate}[(a), itemsep=1pt]
 \item Annotated-train - a small sample of human-annotated data. In this paper, it will be called "small Toxigen". Rows with toxicity scores below 4 or those with annotation disagreements were filtered to maintain data quality. The data set contains 1608 rows after filtering.
 \item Toxigen - the large data set. In this paper, it will be called "large Toxigen". The data set contains 250,951 rows.
\end{enumerate}

\section{Data Preprocessing}
\label{preprocess}
Data preprocessing is an important step before training learning models. Due to the chaotic and noisy nature of language data, this step is even more crucial in the given task. Several methods were performed to clean and normalize the raw data text before fine-tuning BERT, and they are as follows \cite{nguyen-etal-2020-bertweet}:
\begin{enumerate}[itemsep=0.5pt]
\item Excluding posts that are not written in English.
 \item Lowercasing of all tokens to ensure consistent formatting of the text data.
 \item Replacing specific tokens such as "@" mentions and hashtags, and URLs with their corresponding key (e.g. <USER> for a mention).
 \item  Converting emoji tokens to words using the \texttt{demojize} function from the \texttt{emoji} package.
 \item Replacing special characters such as "\textquoteleft{}’\textquoteright{}" and "\textquoteleft{}…\textquoteright{}" with their corresponding ASCII characters.
 \item Separating contractions into their component words and spacing them correctly.
 \item Correctly spacing time expressions such as "p.m." and "a.m.".
 \item Removing any extra whitespace characters to produce the final cleaned and normalized text data.
 \end{enumerate}
 
This text normalization and cleaning process was crucial for preparing the Parler posts for fine-tuning of BERT, as it helped ensure that the text data was in a consistent and relevant format for analysis. 

\section{Computational Approach}
In this work, we are interested in target detection among the hate speech posts. To achieve this task, we suggest a two stages pipeline; (a) Decide whether a post includes hate speech with a designated hate speech detection model. (b) If the answer to the first question is positive, continue and use a hate target classification model.

We fine-tune BERT \cite{DBLP:journals/corr/abs-1810-04805} transformer for each model to the corresponding task and on different data sets. 

\begin{table*}[!ht]
\centering
 \begin{tabular}{|c|c|c|c|c|} 
 \hline
   Model & Accuracy & Recall & Precision & F1 \\ [0.5ex] 
 \hline
 Threshold 3 & 74 & 49 & \textbf{62} & 55 \\  \hline
 Threshold 4 & \textbf{89} & 36 & 40 & 38 \\  \hline
 Weighted Loss Threshold 3 & 75 & \textbf{76} & 59 & \textbf{66} \\  \hline
 Weighted Loss Threshold 4 & 80 & 63 & 30 & 41 \\ \hline
 Weighted Loss Threshold 3 BackTranslation & 76 & 70 & 60 & 65 \\  \hline
 Weighted Loss Threshold 4 BackTranslation & 87 & 34 & 42 & 37 \\ 
 \hline
 \end{tabular}
\caption{Relevant metrics are presented for the proposed Hate Speech Detection models on the Parler Annotated dataset.}
\label{table:model1}
\end{table*}
    
\subsection{Hate Speech Detection}
The first step involves fine-tuning BERT with annotated Parler data. Parler annotated data, as presented above (Section \ref{dataset}), consists of label mean for each example. The original paper \cite{israeli2022free} set a label mean of three and above as a hate speech post. Nonetheless, since posts with label mean above three have been found questionable in their real degree of hate, a threshold larger than three has been set and evaluated as well. The threshold transforms the task into a binary classification one, when posts with a label mean larger than the threshold have been labeled as hate speech, while others as normal. 

Since the skewed nature of the data, we have used a weighted loss, a method to weigh the classes differently. This technique rewards the model more when succeeding in the less common class, and therefore, compensating for the classes' imbalance. Another method for better convergence is the back translation which will be introduced in Section \ref{back_translation}.

\subsection{Target Classifcation}
The second step focuses on fine-tuning Bert on the HateXplain and Dialoconan datasets to identify the target group of hate speech. As it has been mentioned in section \ref{HateXplain}, Four minorities have been chosen; LGBT, Jewish, Muslims, and Black people.
In addition, topic concatenation in each example has been done using the BERTopic model, and will be further discussed in section \ref{BERTopic}

\subsection{Enhancements to Model Performance}
 Both of the models got into an overfit after few epochs. Several methods have been executed in order to prevent this overfit or even delay it and achieve better accuracy in the validation set. These methods include cleaning, normalize, augment the data, and even using the help of other models. 
 
\subsubsection{BERTopic}
\label{BERTopic}
A post's topic can be an related to the hate speech target. Therefore, concatenating the topic of the text, which was identified using BERTopic \cite{grootendorst2022bertopic}, to the input text during the training has been done. BERTopic utilized the BERT embedding representation of the tokens in the posts, followed by clustering on the reduced embedding using the HDBSCAN \cite{mcinnes2017hdbscan} clustering algorithm. The BERTopic model assigned names to the topics based on the most frequent words in each cluster.
One of the difficulties in this approach was determining the optimal parameters for the HDBSCAN model, particularly the min cluster size and sample size, which defined the minimum cluster size and the number of outliers, respectively. To address this challenge, the open-source repository TopicTuner \cite{TopicTuner} was utilized, which is a convenient wrapper for BERTopic models that streamlines the process of finding optimized min cluster size and sample size parameters.

\subsubsection{Back Translation}
\label{back_translation}
Another modification involved the use of back translation, which involves translating the text into five different languages and then back to the original language. This technique enhances the robustness of the models and their ability to handle different language variations and reduce overfitting.
We have implemented the back translation method using MarianNMT, \cite{mariannmt} a library that uses DNN and transformers models to train more than 1000 different pre-trained translation models of various languages. We back translated five different languages: Spanish, German, Russian, French, and Chinese. Out of the five, Russian and Chinese were not used since there were many failing back translations in them compared to the other ones.
Some translations are noisy and contain gibberish. To overcome these issues, we used techniques of text preprocessing (Section \ref{preprocess}) and duplicate words removal.
Using the augmentation created with the back translation technique, we diversify the dataset.

\begin{figure}[!ht]
    \centering
    \includegraphics[height=5.5cm]{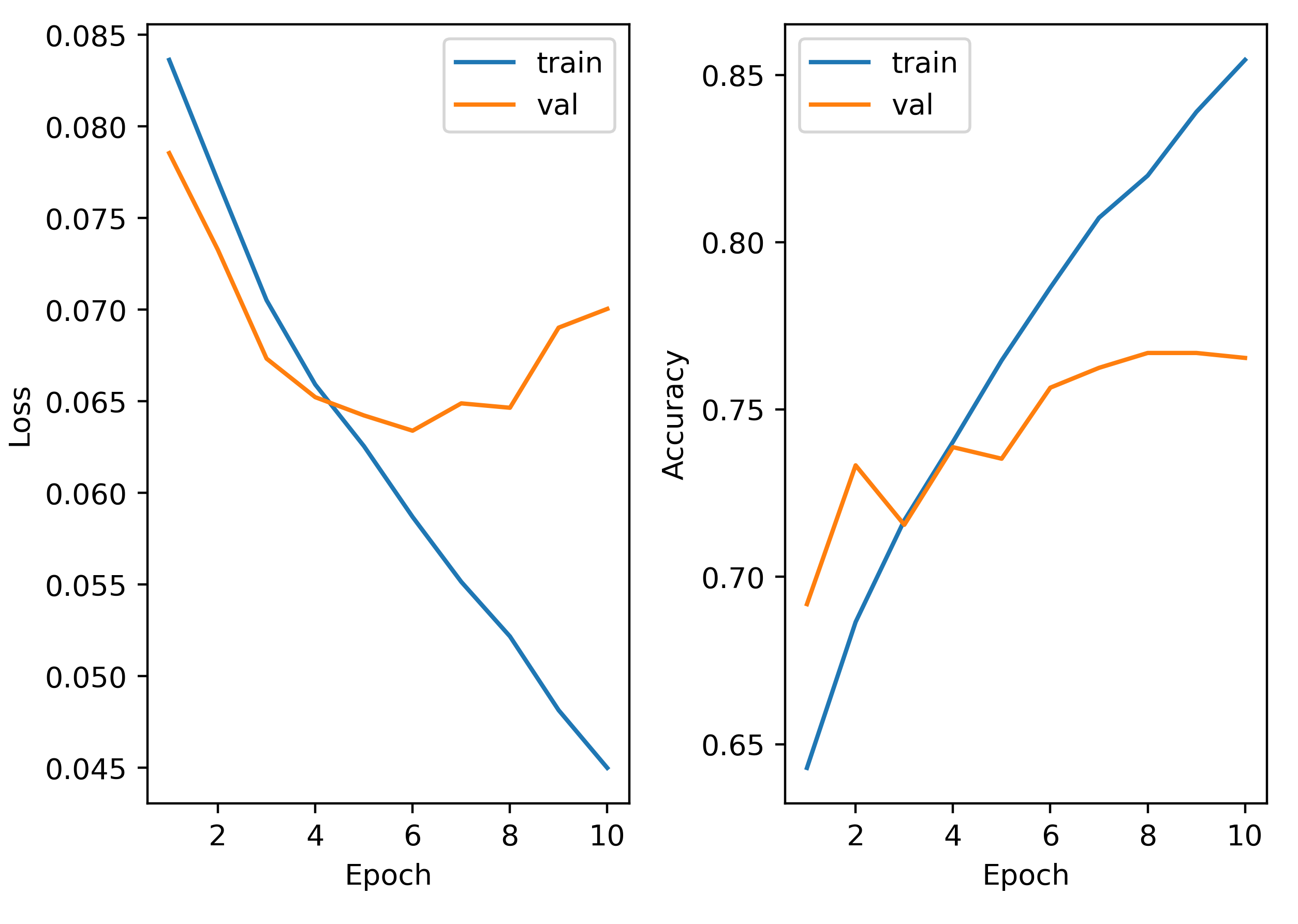}
    \caption{Loss and accuracy trends of Weighted Loss Threshold 3 model, the hate detection best model, are presented.}
    \label{fig:model1_trends}
\end{figure}

\section{Experimental Results}
\subsection{Hate Speech Detection}
The model was trained and executed with a 80:20 train-to-test split in Annotated Parler data \ref{annotated_parler} on a 32GB Tesla V100 GPU, using Pytorch \cite{paszke2019pytorch}. The training was carried out with a batch size of 8, 10 epochs, a cross-entropy loss function, a learning rate of 0.00005, and an Adam optimizer. Numerous parameters permutations such as threshold and weighted loss have been executed (Table \ref{table:model1}). 

The best accuracy was achieved by Threshold 4 model with 89\%, the best recall by Weighted Loss Threshold 3 model with 76\%, the best precision by Threshold 3 model with 62\%, and the best F1 is again the Weighted Loss Threshold 3 model with 66\%. In the hate detection stage, we aim to achieve the lowest false negative, that is, the lowest recall, while still preserving a reasonable F1 score, and therefore, Weighted Loss Threshold 3 is the chosen model. We want to miss a minimum number of hateful posts, even at the cost of a relatively high false positive. This is since an additional "selection" process occurs in the second stage -- the targets detection model classifies any post without one of the predefined targets, as well as non-hateful posts with the "other" class. Learning curves of the chosen model are presented in Fig.\ref{fig:model1_trends}.

\subsection{Target Detection}
\subsubsection{BERTopic}
In the experimental phase of this project, a BERTopic model was utilized to obtain the topic partition of the data. The parameters of the HDBSCAN algorithm were optimized using the TopicTuner repository. The minimum sample size was set to the size of the smallest target in the data set, resulting in the optimal parameters that minimized the occurrence of outliers.
Implementing the optimized HDBSCAN parameters within the BERTopic model resulted in a topic partition of four distinct topics, in addition to one outlier topic(-1). The names of the identified topics are as follows:
\begin{enumerate}[(a), itemsep=1pt]
 \item'0\_nigger\_white\_niggers\_like',
 \item'1\_moslem\_muslim\_sand\_number',
 \item'2\_kike\_jews\_jew\_white',
 \item'3\_faggots\_faggot\_user\_queers'.
 \item'-1\_user\_nigger\_number\_white',
 \end{enumerate}
The topics partition performed by the model revealed a successful detection of targets, aligning with the labels in the original data.

\newcommand{\specialcell}[2][c]{%
\begin{tabular}[#1]{@{}c@{}}#2\end{tabular}}
\begin{table*}[!ht]
    \centering
    \begin{tabular}{|l|l|l|l|l|l|l|l|l|l|l|l|l|l|l|l|l|l|l|l|}
    \hline
        Evaluation dataset & \specialcell{Back\\Translation}&\specialcell{Topic\\in input}& Accuracy & Recall & Precision &F1\\ \hline
        \multirow{4}{*}{Small Toxigen} & \checkmark &\checkmark & 0.5& 0.57 & 0.47&0.47\\  
                                                                     & \checkmark& ~&0.65& \textbf{0.62} & 0.53 & 0.55\\  
                                                                     &        ~    & \checkmark& \textbf{0.71} & 0.57 & \textbf{0.6} &\textbf{0.58}\\ 
                                                                     & ~ & ~& 0.66 & 0.51 & 0.55&0.52\\\hline
                                      
        \multirow{4}{*}{Large Toxigen} & \checkmark &\checkmark &0.53 & 0.57 & 0.47&0.47\\  
                                                                     & \checkmark& &0.63 & \textbf{0.58} & 0.5&0.52\\  
                                                                     &        ~    & \checkmark& \textbf{0.71} & 0.57 & \textbf{0.59}&\textbf{0.57}\\  
                                                                    & ~&~& 0.66 & 0.56 & 0.54&0.54\\\hline
                                                        
        \multirow{4}{*}{Targets Annotated Parler} & \checkmark &\checkmark &0.69 & 0.76 & 0.5&0.56\\  
                                                                     & \checkmark&~ &0.74 & 0.7 & 0.53&0.58\\  
                                                                     &        ~    & \checkmark& \textbf{0.82} & 0.73 & \textbf{0.61}&\textbf{0.66}\\ 
                                                                     &~&~& 0.78 & \textbf{0.79} & 0.59&0.65\\\hline

    \end{tabular}
    \caption{Relevant results of Target detection model, trained on HateXplain dataset.}
\label{table:model2}
\end{table*}

 \subsubsection{Fine-tuning Bert results}
\begin{figure}[!ht]
    \centering
    \includegraphics[height=5.5cm]{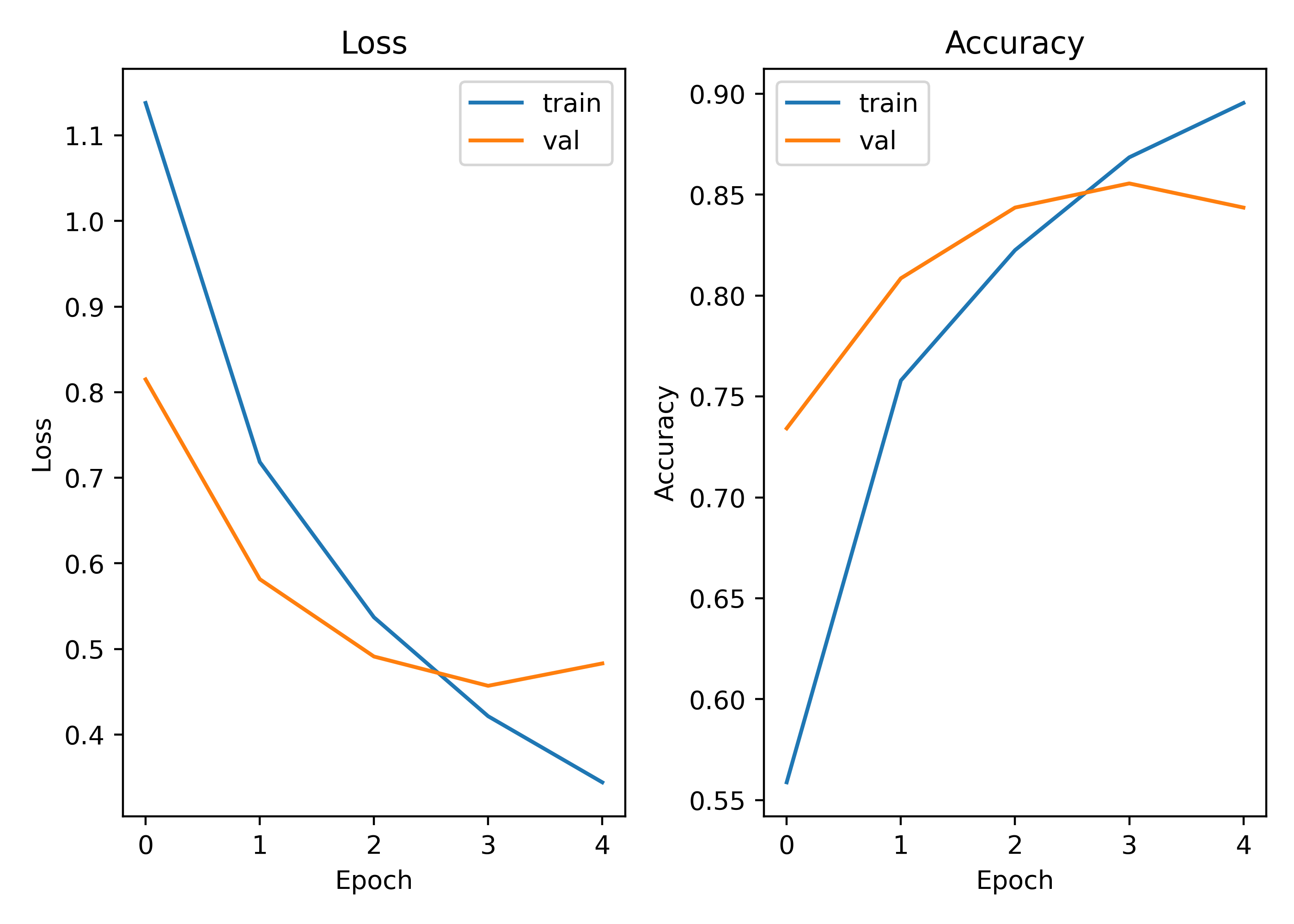}
    \caption{Loss and accuracy trends of the topic concatenated model, trained on HateXplain dataset.}
    \label{fig:model2_acc_loss}
\end{figure}

BERT model was fine-tuned on two training datasets (\ref{HateXplain}, \ref{Dialoconan}), and evaluation was conducted on three separate datasets (\ref{Toxigen}, \ref{target_annotated_parler}). The data was split into a 80:20 ratio.
An attempt was made to merge the training datasets, however, this resulted in a performance decrease. The primary focus was on implementing the following steps to enhance the model's performance: the concatenation of topics names and/or the utilization of back-translation techniques.\\ The accuracy and loss trends of the model with the added topic approach, trained on the HateXplain dataset, are depicted in Fig.\ref{fig:model2_acc_loss}. The model showed signs of overfitting starting from the fourth epoch, which prompted the implementation of early stopping.\\
Using separate evaluation datasets with distributions different from the training data is crucial for accurately assessing the performance of machine learning models and their ability to generalize to new, unseen data. The evaluation results of the models that were trained on the HateXplain dataset (\ref{HateXplain}) are presented in Table \ref{table:model2}.
While the topic concatenation approach consistently showed the best performance, we tested the back-translation approach to enhance the model performance further. However, the results (Table \ref{table:model2}) indicate that the back-translation approach had a diminishing effect on model performance, despite our initial expectation. The use of back-translation introduced noisy or irrelevant data during the translation process, which negatively impacted the model's performance.
The Large Toxigen dataset (\ref{Toxigen}) results (Table \ref{table:model2}) revealed substantial improvement in the topic concatenating model's performance. The improvement was significant and consistent across various metrics. The best improvement observed in the large Toxigen dataset was in the accuracy and precision metrics, highlighting that the major improvement was achieved in reducing false positive errors.

\begin{figure*}[htp]
\label{limefig}
  % Fixed length
  \centering
  \captionsetup[subfigure]{labelformat=empty}
  \subcaptionbox{(A) Hate Speech Model success example is presented.\label{limefig:a}}{\includegraphics[width=2.2in]{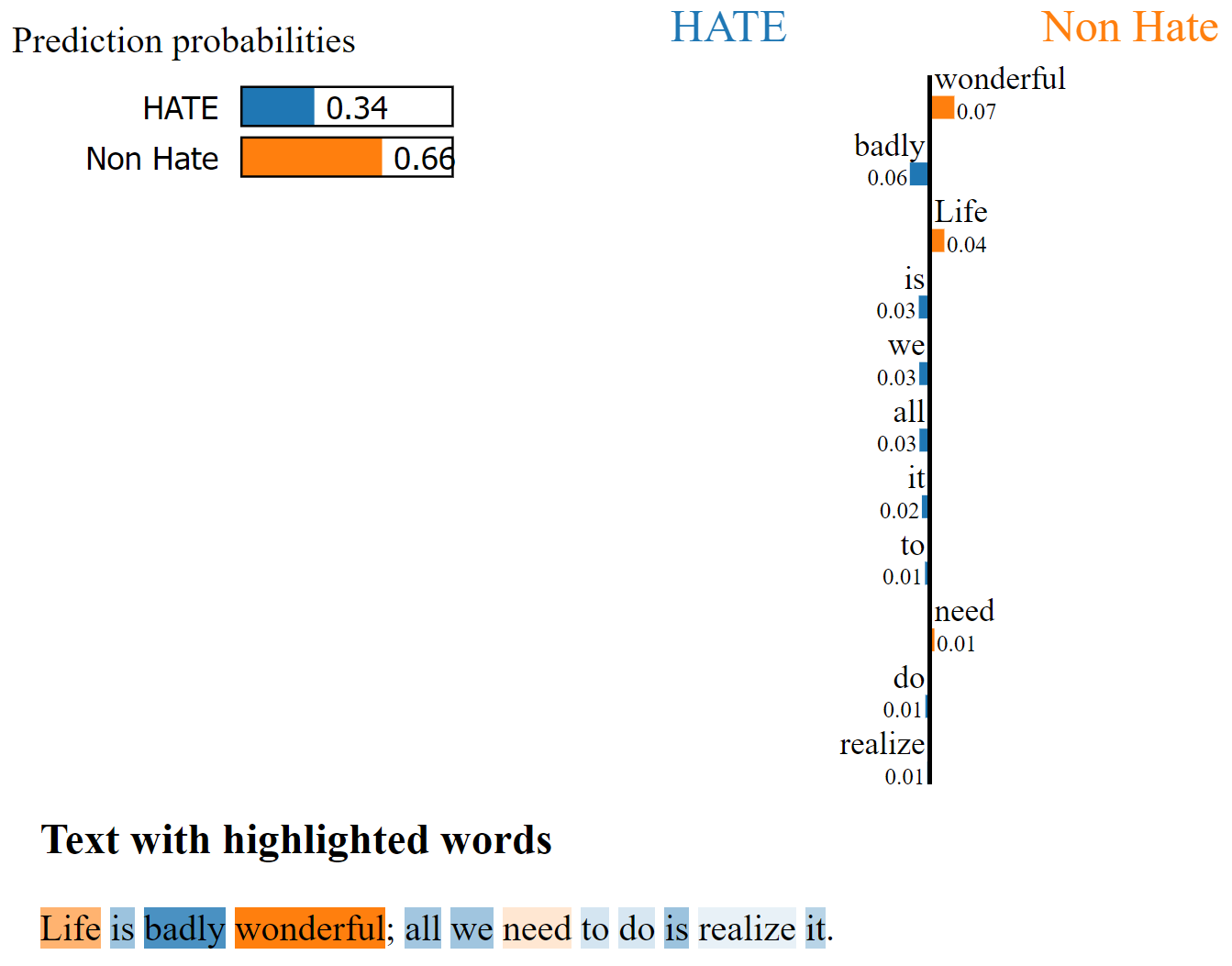}}\hspace{1em}%
  \subcaptionbox{(B) Hate Speech Model failure example is presented.\label{limefig:b}}{\includegraphics[width=2.2in]{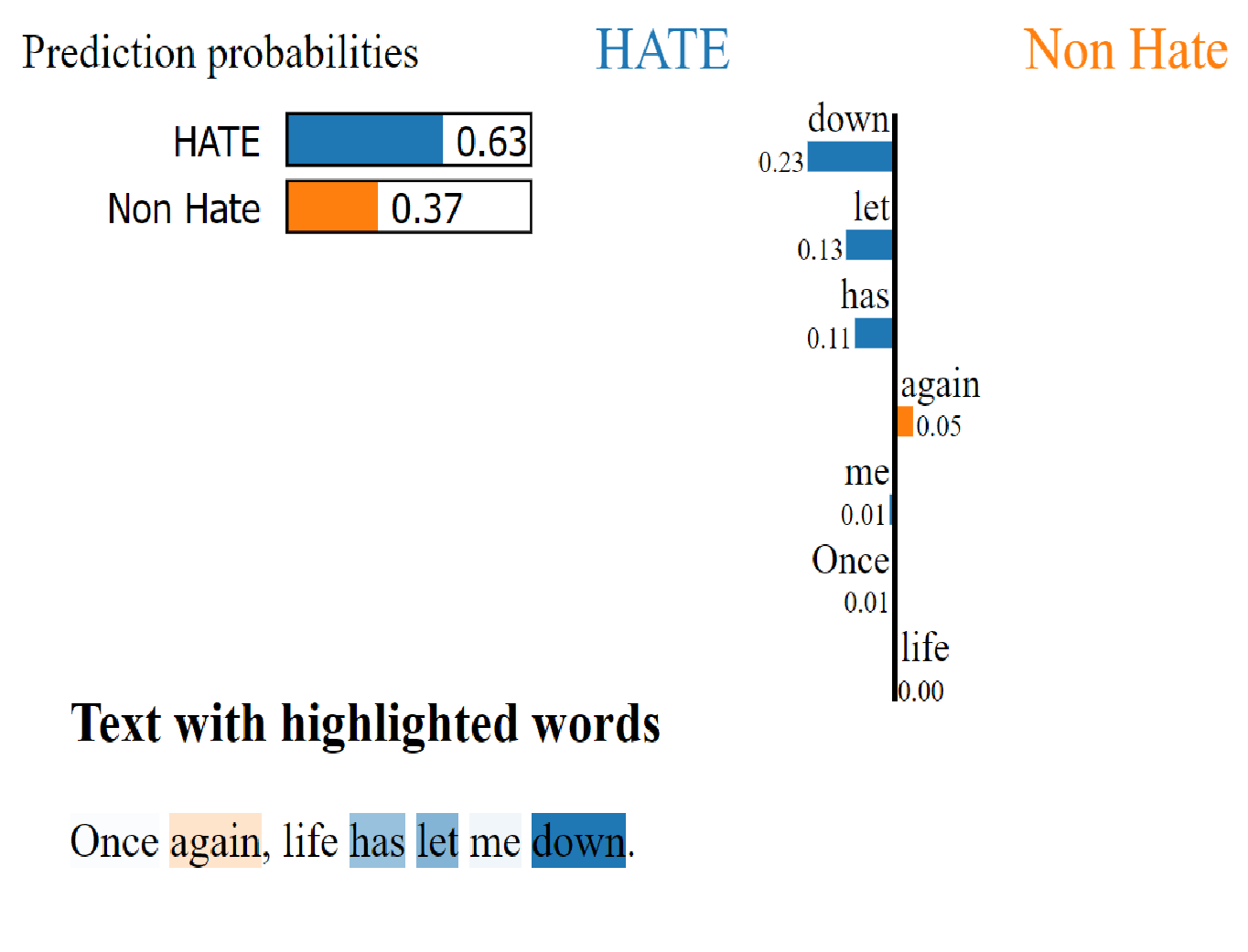}}
  \subcaptionbox{(C) Target Detection Model success example is presented.\label{limefig:c}}{\includegraphics[width=2.2in]{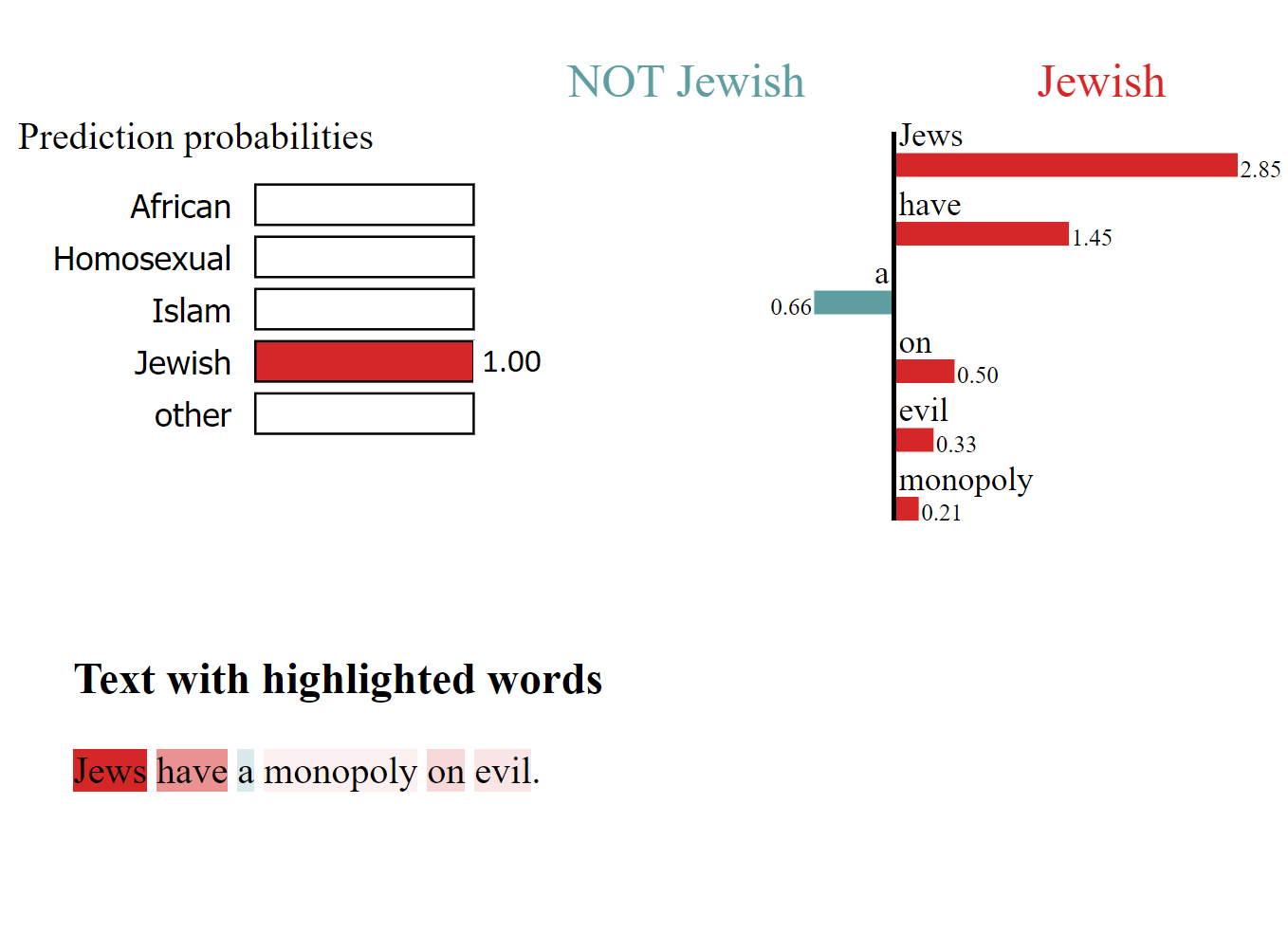}}\hspace{1em}%
  \subcaptionbox{(D) Target Detection Model failure example is presented.\label{limefig:d}}{\includegraphics[width=2.2in]{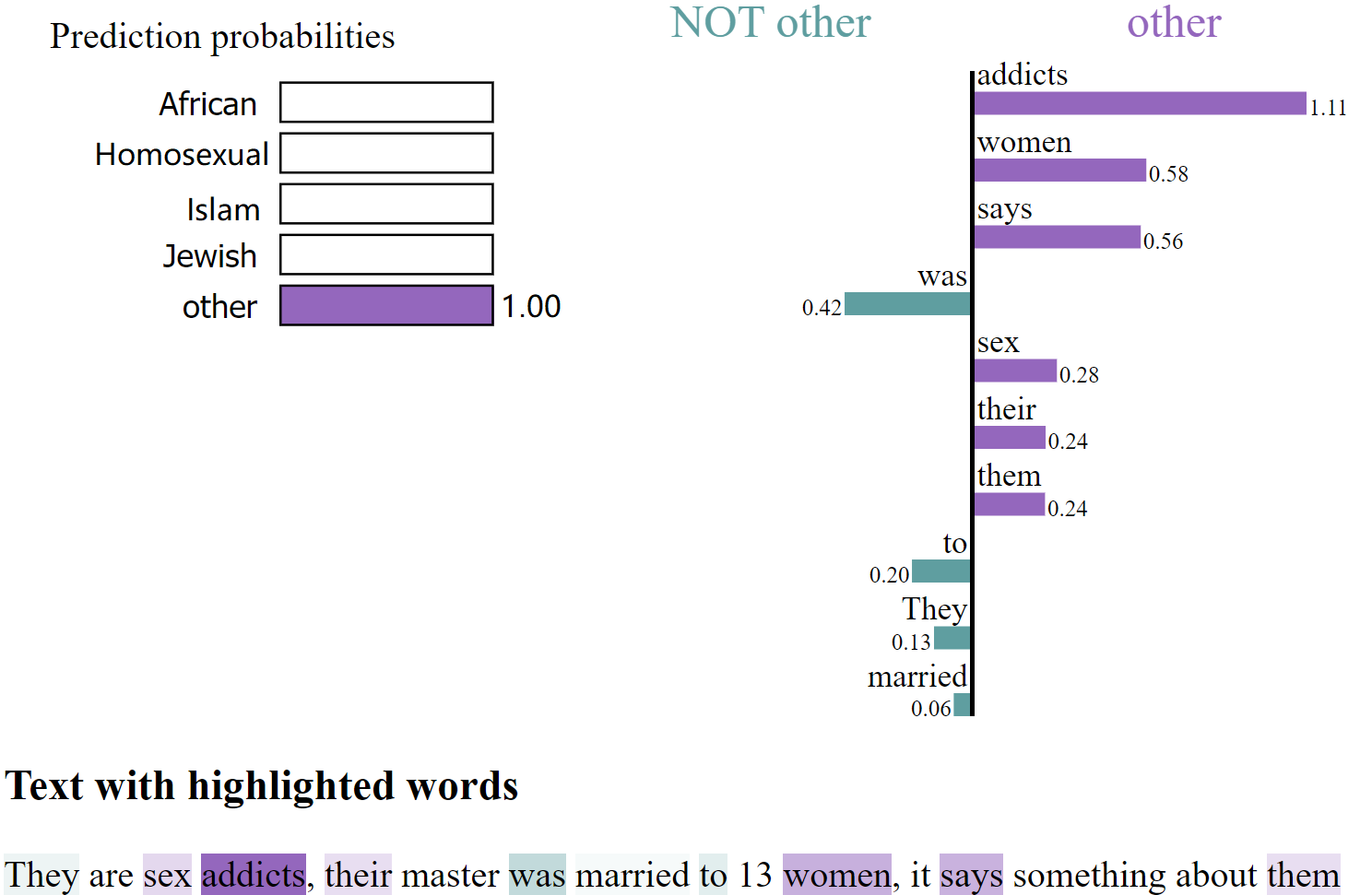}}
  \caption{Examining samples using LIME in both the hate speech classification model and the target classification model.}
\end{figure*}

\section{Model Explainability}
Valid evaluation is a crucial step in assessing of every model. During an evaluation, the model's predictions are compared against predetermined labels of the chosen data set, however, the given data set doesn't have any.
Evaluation on other data sets (Table \ref{table:model2}) can reflect performance to some degree, but to raise the credibility of our suggested models, explainability techniques have been used. Explainability attempts to explain the reason behind a decision, and hence, enabling us to ensure the model's focus is on relevant features only.

LIME \cite{lime} algorithm was implemented on the models' predictions to study the connections
between inputs and outputs. Success and failure examples are presented for each model:
\begin{enumerate}[(A), itemsep=1pt]
 \item Hate Speech Model Success (Fig.\ref{limefig:a}) -- "Life is badly wonderful: all we need to do is realize it." The emphasized positive words are "Life" and "wonderful" while a negative word is "badly" as expected. The model understood the broad context despite the mix of positive-negative words, and the post was classified as a non-hate.
 \item Hate Speech Model Failure (Fig.\ref{limefig:b}) -- "Once again, life has let me down." The negative words are "let" and "down" while a positive word is "again" as expected. This example was wrongly classified as a hate post. Negative words are presented but there is no minority in it, and thus, no hate speech.
 \item Target Detection Success (Fig.\ref{limefig:c}) -- "Jews have a monopoly on evil." The emphasized Jewish contribution word is "Jews" as expected, and model indeed classified the target as Jewish.
 \item Target Detection Failure (Fig.\ref{limefig:d}) –- "They are sex addics, their master was married to 13 women, it says something about them." The most impactful words are "addicts" and "women". Both are groups in the "other" class and this is also the model's decision, however, the model didn't recognize the broad context in which the true target is Muslims.
 \end{enumerate}
 
\section{Conclusions and Future Work}
In this paper, we present a pipeline for extraction of hate speech targets in Parler data set. Two finetuned BERT models have been trained, and various methods for performance enhancement have been tested. Targets detection model was trained on different data sets than Parler. To raise the confidence of our final Parler targets distribution, we present a new targets Parler annotated data set -- TAP (Section \ref{target_annotated_parler}). This data set is part of Parler, and thus, its evaluation is more representative. In addition, explainability of the models has been done to investigate the models' performance further and hence, the final targets distribution reliability.

\begin{figure}
    \centering
    \includegraphics[height=5.5cm]{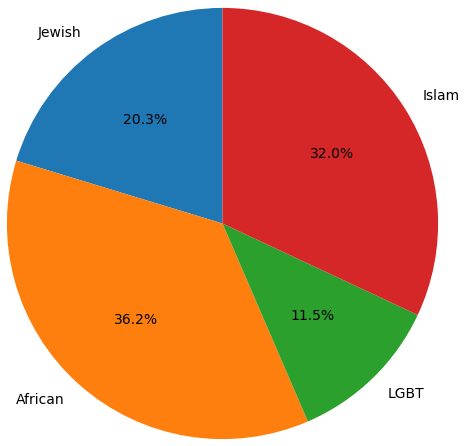}
    \caption{Hate Speech Targets distribution out of 824,848 Parler posts.}
    \label{fig:parler_dist}
\end{figure}

17\% out of approximately 800,000 Parler posts are hate speech posts, and most of the targets are Africans (Fig.\ref{fig:parler_dist}). This result leads to ethical questions and may have a link to the origin of Parler users and perhaps even the time in which the data was scraped. The final distribution is similar to TAP (Section \ref{target_annotated_parler}) distribution.

Future work includes completing parsing the whole Parler data set, continue improving the models' performance, whether with additional data or optimization methods and perhaps use unsupervised techniques on the big Parler data set.

\section*{Acknowledgements}

\bibliography{utils/custom}
\bibliographystyle{utils/acl_natbib}
\end{document}